\pdfoutput=1

\documentclass[11pt]{article}

\usepackage{EMNLP2023}

\usepackage{times}
\usepackage{latexsym}

\usepackage[T1]{fontenc}

\usepackage[utf8]{inputenc}

\usepackage{microtype}

\usepackage{inconsolata}

\usepackage{graphicx}
\usepackage{booktabs}
\usepackage{comment}
\usepackage{multirow}

\title{Summarization-based Data Augmentation for Document Classification}

\author{Yueguan Wang \\
  The University of Tokyo \\
  \texttt{etsurin@iis.u-tokyo.ac.jp} \\\And
  Naoki Yoshinaga \\
  Institute of Industrial Science,\\
The University of Tokyo \\
  \texttt{ynaga@iis.u-tokyo.ac.jp} \\}

\newcommand{\yn}[1]{\textcolor{black}{#1}}
\newcommand{\et}[1]{\textcolor{black}{#1}}

\begin{document}
\maketitle
\begin{abstract}
Despite the prevalence of pretrained language models in natural language understanding tasks, understanding lengthy text such as document is still challenging due to the data sparseness problem.
Inspired by that humans develop their ability of understanding lengthy text from reading shorter text,
we propose a simple yet effective summarization-based data augmentation, SUMMaug, for document classification.
We first obtain easy-to-learn examples for the target document classification task by summarizing the input of the original training examples, while optionally merging the original labels to conform to the summarized input. We then use the generated pseudo examples to perform curriculum learning.
Experimental results on two datasets confirmed the advantage of our method compared to existing baseline methods in terms of robustness and accuracy.
We release our code and data at \url{https://github.com/etsurin/summaug}.
\end{abstract}

\section{Introduction}

Although the pretrained language models~\cite{devlin-etal-2019-bert,liu2019roberta,debarta2020} have boosted the accuracy of various natural language understanding tasks, the accuracy is still limited for complex tasks
with lengthy input~\cite{lin-etal-2023-linear} and fine-grained output~\cite{liu-etal-2021-effective}, such as document classification. These tasks require models to find a mapping between diverse input and output, which models are more likely to suffer from the data sparseness problem.


To address the data sparseness problem, researchers have studied data augmentation
for text classification tasks. A basic approach is to generate pseudo training examples from gold examples by perturbing the inputs; those perturbation include back-and-forth translation~\cite{shleifer2019low} and minor editing of input text~\cite{wei-zou-2019-eda,karimi-etal-2021-aeda-easier} or its hidden representations~\cite{chen-etal-2020-mixtext, chen-etal-2022-doublemix, wu-etal-2022-text}. These methods basically echo the information in the original training data, 
\yn{which will not help much the model learn to read lengthy inputs.} 


In this study, to effectively develop the model's ability to comprehend the content in document classification, we propose a simple yet effective summarization-based data augmentation, SUMMaug, to generate pseudo, abstractive training examples for document classification. Specifically, we apply text summarization to the input of gold examples in document classification task to obtain abstractive, easy-to-read examples, and merge fine-grained target labels as needed so that the labels conforms to the summarized input. Motivated by that we humans gradually develop the ability of understanding lengthy text from reading shorter text, we use the generated examples in the context of curriculum learning (surveyed in~\cite{soviany2022curriculum}), namely, curriculum fine-tuning.


We compare our method to a baseline data augmentation~\cite{karimi-etal-2021-aeda-easier} on two versions of IMDb dataset with a different number of target labels.
Experimental results confirm that curriculum fine-tuning with SUMMaug outperforms baseline methods on both accuracy and robustness. 

\begin{figure}
    \centering
    \includegraphics[width=\linewidth]{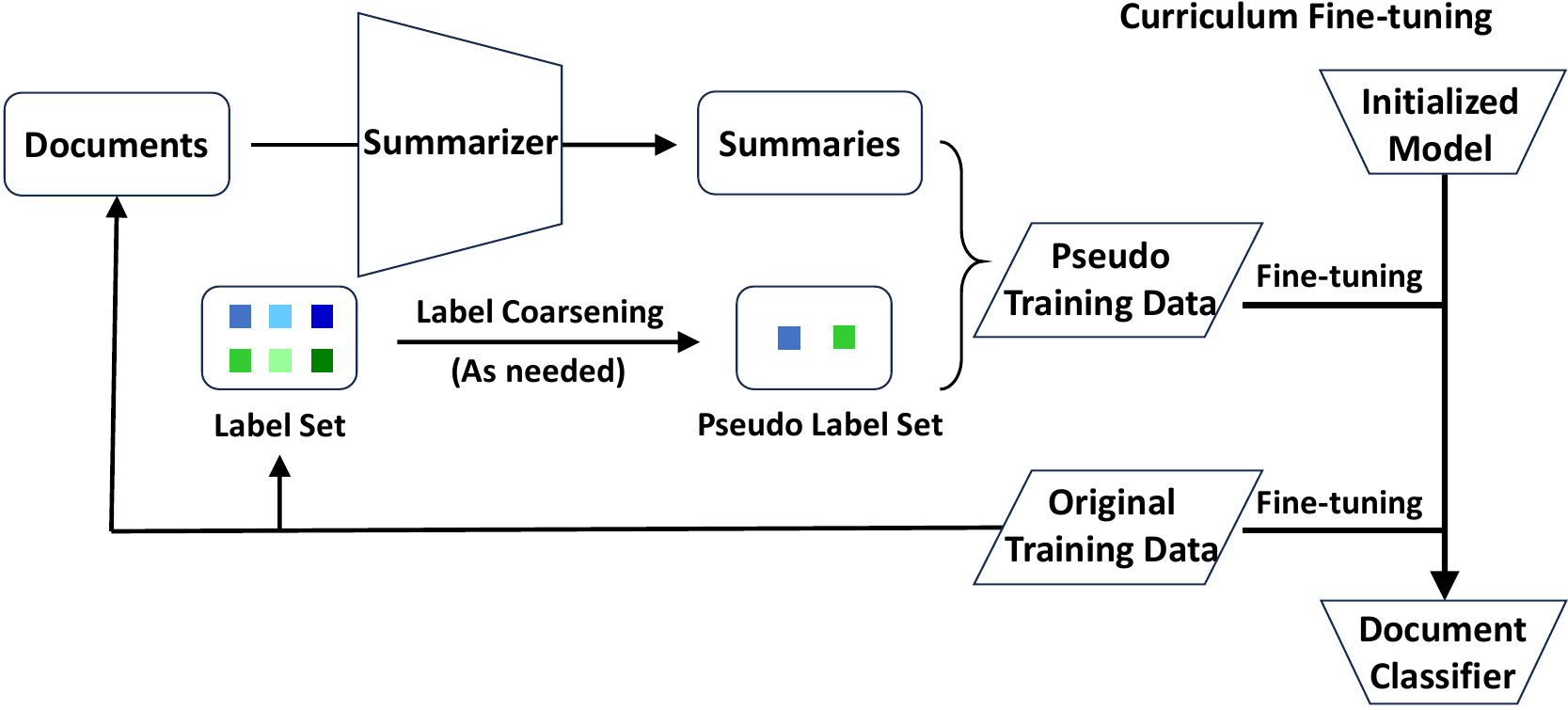}    \caption{Curriculum fine-tuning for document classification using SUMMaug data augmentation: \yn{prior to the normal finetuning, it fine-tunes a model with easy-to-learn examples obtained by summarizing the original training examples}.}
    \label{flow}
\end{figure}


\section{Related Work}
In this section, we first review existing neural models for document classification, and next introduce existing data augmentation methods for text classification. We then mention other attempts to leverage summarization for text classification.

\paragraph{Document Classification}
In the literature, researchers 
\yn{explore}
a better neural architecture to comprehend the lengthy content in document classification; examples include a graph neural network~\cite{zhang-zhang-2020-text,zhang-etal-2022-hierarchical}
and 
\yn{a convolutional attention network}~\cite{liu-etal-2021-effective}. Recently, Transformer~\cite{NIPS2017_3f5ee243}-based models have been revisited~\cite{dai-etal-2022-revisiting} and reported to outperform the task-specific networks. Since our work is model-agnostic and orthogonal to the model architecture, we adopt RoBERTa~\cite{liu2019roberta}, a Transformer-based pre-trained model, as the target of evaluation.


\paragraph{Data Augmentation for Text Classification}
To address the data sparseness problem in text classification, researchers employ data augmentation, which generates pseudo training examples from the training examples. \citet{shleifer2019low} leverages back-and-forth translation 
to paraphrase the inputs of training examples.
Through translating the inputs into another language and then translating  the resulting translation back to the source language, they obtain the input that are written in different ways but will have the same meanings conforming to the corresponding target labels. \citet{DBLP:conf/iclr/XieWLLNJN17} perturb the input by deleting and inserting words and replacing words with their synonyms. \citet{karimi-etal-2021-aeda-easier} propose a simple but more effective perturbation that randomly inserts punctuation marks. Rather than directly perturbing the input of training examples, some studies add noises in their continuous representations~\cite{chen-etal-2020-mixtext,chen-etal-2022-doublemix,wu-etal-2022-text}. \et{However, these method predominantly echo existing training data, providing minimal assistance in understanding lengthy texts. }


\begin{table*}
\centering
\begin{tabular}{p{0.97\textwidth}}
\toprule
\small{I am Anthony Park, Glenn Park is my father. First off I want to say that the story behind this movie and the creation of the Amber Alert system is a good one. However \textcolor{red}{the movie itself was poorly made and the acting was terrible}. The major problem I had with the movie involved the second half with Nichole Timmons and father Glenn Park. \textcolor{red}{The events surrounding that part of the story were not entirely correct.} My father was suffering from psychological disorders at the time and picked up Nichole without any intent to harm her at all. He loved her like a daughter and was under the mindset that he was rescuing her from some sort of harm or neglect that he likely believed was coming from her mother who paid little attention to her over the 3 plus years that my father took care of her and summarily raised her so her mother could frolic about. The movie depicted my father in a manner that he was going to harm her in some way shape or form. The funny thing is that Nichole had spent many nights sometimes consecutively at my fathers place while Sharon would be working or doing whatever she was doing. The reason that my father was originally thought to be violent was because he had items that could be conceived to be weapons on his truck. My father was a landscaper. The items they deemed to be weapons were landscaping tools that he kept in his truck all the time for work. \textcolor{red}{My recommendation is take this movie with a grain of salt, it is a good story and based on true events} however the details of the movie (at least the Nichole Timmons - Glenn Park portion) are largely inaccurate and depict the failure of the director to discover the truth in telling the story. The funny thing is, that if the director would have interviewed any of Sharon's friends who knew the situation they would have stated exactly what I have posted here.}
 \\\midrule
\small{The movie itself was poorly made and the acting was terrible. The events surrounding that part of the story were not entirely correct. My recommendation is take this movie with a grain of salt, it is a good story and based on true events.}
\\\bottomrule
\end{tabular}
\caption{\label{example} An example of original text an generated summary on IMDb dataset. The first row is the original text while the second row is the generated summary. \textcolor{red}{Red} text are counterparts of summary in the original text. }
\end{table*}

\paragraph{Use of Summarization in Text Classification}
\citet{li-zhou-2020-connecting} and \citet{hartl-kruschwitz-2022-applying} utilize automatically generated summaries to retrieve fact for fake news detection. Whereas this approach uses summaries to retrieve knowledge for classification, our approach leverages summaries in training as easy-to-learn examples, which does not assume costly summarization in inference.

\section{SUMMaug}
Document classification requires a model to comprehend lengthy text \yn{with dozens of sentences,} which is even difficult for humans, \yn{especially, children and second-language learners.} Then, how do we humans develop an ability to comprehend lengthy text? In school, starting from reading short, concise text, we gradually read longer text.

In this study, we develop a summarization-based data augmentation \yn{method} for document classification, SUMMaug, and use \yn{it to generate} pseudo, abstractive training examples from gold examples to perform curriculum learning in document classification.


\subsection{Summarization-based data augmentation}
\label{sec:idea}
In SUMMaug, a summarization model $M$ is used to generate pseduo, easy-to-learn examples for document classification. \yn{In this study, we apply an off-the-shelf summarization model, $M$, to} each training pair $\{x,y\}$, where $x$ denotes the document and $y$ denotes the label, and \yn{then obtain} a concise summary \yn{of $x$,} namely, $\hat{x} = M(x)$. 

An issue here is how to determine the label for the generated concise summary, $\hat{x}$. Since the summarization abstracts away detailed information for classification, the original target label $y$ can be inappropriate especially when the target labels are fine-grained.
We thus define a map function $f$ to merge the fine-grained categories into a coarse-grained label group, and obtain the augmented training pair is $\{\hat{x},f(y)\}$, as shown in Figure~\ref{flow}.

\paragraph{On summarization model} To summarize diverse text handled in document classification, we assume an off-the-shelf summarization model that can handle documents with diverse topics. In this study, we choose an off-the-shelf BART~\cite{lewis-etal-2020-bart}-based summarization model fine-tuned on CNN-Dailymail~\cite{hermann2015teaching} dataset as an implementation of $M$,\footnote{\url{https://huggingface.co/facebook/bart-large-cnn}}
since the writing style of news reports is suitable for most of the text in daily life. 
We should mention that the CNN-Dailymail dataset contains mostly extractive summaries, and the resulting summarization model will be less likely to suffer from hallucinations~\cite{maynez-etal-2020-faithfulness} \yn{that have been} reported for a summarization model trained on abstractive summarization datasets such as XSum~\cite{narayan-etal-2018-dont}.

Table~\ref{example} exemplifies a summary generated for IMDb datasets. While the \yn{original input} (review) exhibits a mild negative sentiment, its compression into a summary intensifies this sentiment. This observation underscores the imperative to categorize labels of augmented data into coarser groups.
 
\subsection{Learning a Classifier with augmented data}
In the literature of data augmentation, the models are basically trained with the original and augmented training data, since both data are related to the target task. In our settings, however, the labels will be merged into fewer labels so that the labels conform to the generated summaries. We thus consider the following two strategies to utilize the pseudo abstractive training data.

\begin{description}
\item[Mixed fine-tuning] We combine the original and pseudo training data to fine-tune a pre-trained model for classification. In this setting, we do not collapse labels, namely, $f(y)=y$.
\item[Curriculum fine-tuning] We first finetune a pre-trained model on the pseudo training data, and then finetune a pre-trained model on the original training data. This strategy is inspired by curriculum learning~\cite{curr}. In this setting, we collapse labels as needed. \yn{When we collapse labels, we discard parameters for the collapsed labels in the fine-tuning with the original examples.}
\end{description}
In \yn{the following} experiments, we compare two strategies for datasets with different numbers of labels.

\section{Experiments}
We conduct experiments on two datasets to evaluate our method, thus demonstrating that: \et{(1) our method shows better accuracy and robustness compared with baseline methods in both general setting and low-resource settings; and (2) curriculum fine-tuning plays an important role in achieving improvements. }


\subsection{Dataset}
We use two versions of large-scale movie reviews dataset IMDb for evaluation. One contains 50,000 movie reviews with a positive or negative label~\cite{maas-etal-2011-learning}, while the other involves 10 different labels from rating 1 to 10. 
For the IMDB-2 dataset, we split 10\% of the training data for validation. For the IMDb-10 dataset, the same \yn{splitting} as \citet{adhikari2019docbert} is used.
The detailed information of the two datasets is shown in Table~\ref{datasets}.

\begin{table}[t]
\centering
\small
\tabcolsep4.5pt
\begin{tabular}{lcccccc}
\toprule
Dataset &  train & val & test & $C$ & $L$ & $L_{M}$\\\midrule
IMDb-2 & 22500 & 2500 & 25000 & 2 & 279.5 & 51.3\\
IMDb-10  & 108670 & 13432 & 13567 & 10 & 394.2 & 50.2\\
\bottomrule

\end{tabular}
\caption{\label{datasets} Details of \yn{the IMDb} datasets: $C$ denotes the number of classes. $L$ and $L_M$ denote the average length of the inputs and the generated summaries, respectively.}
\end{table}

\subsection{Methods}
We use the following three models for evaluation. All models are based on RoBERTa~\cite{liu2019roberta} with a classification layer. 

\paragraph{RoBERTa}
We finetune a pre-trained RoBERTa\footnote{\url{https://huggingface.co/roberta-large}} on the original training data as a baseline. 
 
\paragraph{RoBERTa + AEDA}
We use AEDA \cite{karimi-etal-2021-aeda-easier}, a strong data augmentation method for text classification as another baseline. \et{We apply AEDA\footnote{\url{https://github.com/akkarimi/aeda_nlp}} to the original documents}, \yn{and then fine-tune a RoBERTa model on the augmented data and original data.}

\paragraph{RoBERTa + SUMMaug}
We use BART-based summarizer trained on CNN-Dailymail to generate concise summaries, 
and \yn{fine-tune a} RoBERTa model on the augmented data and original data. 

To evaluate the performance of our method in low resource settings, we randomly select 200 and 1500 samples from the two datasets and train a model on these sub datasets. \et{However, on the IMDb-10 dataset, we observe that \yn{all} models diverge and perform randomly when training data is reduced to 200, likely due to the challenges of fine-grained classification with rather limited training data; we \yn{thereby} do not report the results. }

In order to reveal the effectiveness of curriculum fine-tuning, we apply curriculum fine-tuning not only to SUMMaug but also to AEDA\@. On the IMDb-10 dataset, we map the labels of the augmented data into coarse-grained ones, as mentioned in \S~\ref{sec:idea}. Specifically, labels between 0-4 are mapped into 0 (negative) while labels between 5-9 are mapped into 1 (positive).



\begin{table}
\centering
\small
\tabcolsep2.7pt
\begin{tabular}{lccc}
\toprule
\multirow{2}{*}{Model} & \multicolumn{3}{c}{The size of training data} \\
\cmidrule(lr){2-4}
& 200 & 1500 & all\\ \midrule
RoBERTa & 92.19$_{1.21}$ & 94.21$_{0.62}$  & 94.63$_{0.56}$ \\
+ AEDA (mixed) & 90.91$_{1.44}$ & 94.43$_{0.49}$ & 94.75$_{0.66}$ \\
\midrule
+ AEDA (curriculum) & \textbf{93.59}$_{1.16}$ & 94.26$_{0.74}$ & \textbf{95.56}$_{0.12}$ \\
+ SUMMaug (mixed) & 92.94$_{0.99}$ & 94.61$_{0.64}$ & 94.85$_{0.62}$ \\
+ SUMMaug (curriculum)& 93.36$_{0.97}$ & \textbf{94.77}$_{0.28}$ & 95.45$_{0.17}$ \\
\bottomrule

\end{tabular}
\caption{\label{results} Classification accuracy$_\textrm{stdev.}$ (\%) on IMDb-2: mixed and curriculum denotes mixed and curriculum fine-tuning. All the results are averages over five runs. The best \yn{results are} marked as \textbf{bold}.}
\end{table}

\subsection{Implementation Details}
We set the model's hyperparameters as follows. For experiments on the IMDb-2 dataset, batch size is set to 64 and learning rate is set to 1e-5. For experiments on the IMDb-10 dataset, following \citet{adhikari2019docbert}, batch size is set to 16, with learning rate set to 2e-5. \et{Detailed information of training epochs can be found at \hyperref[appen]{Appendix A\@.}} All the experiments were conducted on four NVIDIA Quadro P6000 GPUs with 24GB memory. 

The final model for evaluation is selected \yn{on the basis of} the performance on validation set. To eliminate the effect of random factors, we report the average accuracy over five runs. 
\section{Results}

\begin{table}
\centering
\small
\begin{tabular}{lll}
\toprule 
\multirow{2}{*}{Model} & \multicolumn{2}{c}{The size of training data} \\
\cmidrule(lr){2-3} 
& \multicolumn{1}{c}{1500} & \multicolumn{1}{c}{all} \\
\midrule RoBERTa & 39.99$_{8.46}$ & 56.58$_{0.34}$  \\
+ AEDA (mixed) & 36.58$_{10.64}$ & 51.23$_{14.39}$  \\
\midrule
+ AEDA (curriculum)& 41.77$_{3.01}$ & 56.63$_{1.65}$  \\
+ SUMMaug (mixed) & 40.65$_{2.71}$ & 55.81$_{2.00}$ \\
+ SUMMaug (curriculum) & \textbf{42.14}$_{1.48}$ & \textbf{57.55}$_{0.29}$ \\
\bottomrule
\end{tabular}
\caption{\label{results10} \yn{Classification accuracy$_\textrm{stdev.}$ (\%) on IMDb-10. All the results are averages over five runs. The notations follow Table~\ref{results}. }}
\end{table}

Tables~\ref{results} and \ref{results10} list the results of baseline methods and our proposed method. Our method outperforms baseline methods in all experimental settings. We additionally confirm on both datasets that our data augmentation is effective even when the training data size is small.

\paragraph{How robustly does SUMMaug work?}
SUMMaug \et{achieves higher} classification accuracy across datasets \et{while improving or maintaining robustness (\yn{low standard deviations})}, whereas the original AEDA, namely AEDA (mixed), reduces the accuracy on IMDb-2 when 200 training examples are used, and \et{it leads to unstable results} on IMDb-10 dataset. 

\paragraph{Is curriculum fine-tuning effective?}
We use mixed fine-tuning with SUMMaug and curriculum fine-tuning with AEDA\@. We observe that \et{under mixed fine-tuning method, the data augmented by SUMMaug exhibited less improvements and even turns to be harmful on the IMDb-10 dataset.}
Conversely, it turns out that curriculum learning helps the AEDA method achieve further improvements in some cases while addressing the low robustness issue. However, curriculum learning with AEDA does not consistently enhance results because the AEDA augmented data retains the same information as the original data, which offers limited benefits in improving text comprehension.

\begin{table}
\centering
\small
\begin{tabular}{lcc}
\toprule $N$ & $f$ & Accuracy$_\textrm{stdev.}$ \\
\midrule 2 & [0,0,0,0,0,1,1,1,1,1] & 57.55$_{0.29}$  \\
3 & [0,0,0,1,1,1,1,2,2,2] & 57.47$_{0.20}$  \\
4 & [0,0,0,1,1,2,2,3,3,3] & 57.66$_{0.31}$  \\
5 & [0,0,1,1,2,2,3,3,4,4] & 57.32$_{0.60}$ \\
10 & [0,1,2,3,4,5,6,7,8,9] & 57.20$_{0.56}$ \\
\bottomrule
\end{tabular}
\caption{\label{coarse} \et{Classification accuracy$_\textrm{stdev.}$ (\%) on IMDb-10 with different label coarsening function under SUMMaug (curriculum) method. $N$ denotes the number of merged label groups while $f$ shows how the original label 0-9 is mapped into coarse-grained label. All the results are averages over five runs.}}
\end{table}

\paragraph{How label coarsening affects accuracy?}
\et{Table~\ref{coarse} shows the results of SUMMaug (curriculum) under different map function $f$. The accuracy is comparable when $N \leq 4$, while there's a noticeable decline in accuracy, accompanied by decreased stability when label coarsening is insufficient or not adopted. This is probably because the summaries can filter out \yn{detailed content, which is}
essential for fine-grained classification. On the other hand, unlike mixed fine-tuning, in which potentially noisy augmented data is used throughout the training process, in the curriculum fine-tuning, the effect of noise diminishes after model turns to train on the original data. Consequently, it can still achieve improvement even without label coarsening.}

\section{Conclusion and Future Work}

This \yn{study} explores a novel application of a summarization model and proposes a simple yet effective data-augmentation method, SUMMaug, for document classification. It performs curriculum learning-style fine-tuning to first train a model on concise summaries prior to the fine-tuning on the original training data. This mirrors the human process of mastering lengthy text comprehension, through gradual exposure to longer text. Experimental results on two document classification datasets confirm that SUMMaug enhances both accuracy and training stability compared to the baseline data augmentation method. \et{Meanwhile, our method shows effective in low-resource settings.}

\et{The future work will focus on searching for the optimal mapping function $f$ and exploring the effect of different summarization models. We will also apply SUMMaug to other document classification tasks of various domains.}

\section*{Limitations}
One of the drawbacks of this \et{study} is that we do not \et{consider the label coarsening function $f$ as a hyper-parameter 
 and just choose the simplest one for experiments. 
 The effect of label coarsening function on accuracy is still insufficiently explored.} For the datasets, despite the different numbers of labels, the documents used are originally from the same kind of domain, which is not convincing enough to show that SUMMaug is robust across diverse classification tasks in different domains. 

\section*{Acknowledgements}
This work was partially supported by the special fund of Institute of Industrial Science, The University of Tokyo, and by JSPS KAKENHI Grant Number JP21H03494.

\bibliography{anthology,custom}
\bibliographystyle{acl_natbib}

\appendix
\section{Detailed training epochs}

\et{Table~\ref{epoch} shows detailed training epochs in our experiments. We select training epochs based on the accuracy on the validation set. 
}

\label{appen}
\begin{table}
\centering
\small
\begin{tabular}{lccc}
\toprule
\multirow{2}{*}{Dataset (fine-tuning method)} & \multicolumn{3}{c}{The size of training data} \\
\cmidrule(lr){2-4}
& 200 & 1500 & all\\ \midrule
IMDb-2 (w/o data augmentation) & 70 & 18 & 2 \\
IMDb-2 (mixed) & 70 & 18 & 2 \\
IMDb-2 (curriculum) & 70/70 & 18/18 & 2/2 \\
IMDb-10 (w/o data augmentation) & - & 20 & 4 \\
IMDb-10 (mixed) & - & 20 & 4 \\
IMDb-10 (curriculum)& - & 5/20 & 2/6 \\
\bottomrule

\end{tabular}
\caption{\label{epoch} Detailed training epochs in our experiments. For curriculum fine-tuning method, $x/y$ denotes that model is trained $x$ epochs on augmented data and then $y$ epochs on original data. }
\end{table}
\end{document}